\pdfoutput=1

\documentclass[11pt]{article}

\usepackage{ACL2023}

\usepackage{times}
\usepackage{latexsym}
\usepackage{graphicx}
\usepackage{booktabs}
\usepackage{subfigure}
\usepackage{amsmath}

\usepackage[T1]{fontenc}

\usepackage[utf8]{inputenc}

\usepackage{microtype}

\usepackage{inconsolata}

\title{How do Large Language Models Learn In-Context? Query and Key Matrices of In-Context Heads are Two Towers for Metric Learning}

\author{Zeping Yu \quad Sophia Ananiadou\\
  Department of Computer Science, National Centre for Text Mining \\
  The University of Manchester  \\
  \texttt{\{zeping.yu@postgrad. sophia.ananiadou@\}manchester.ac.uk}}

\begin{document}
\maketitle
\begin{abstract}
We investigate the mechanism of in-context learning (ICL) on sentence classification tasks with semantically-unrelated labels ("foo"/"bar"). We find intervening in only 1\% heads (named "in-context heads") significantly affects ICL accuracy from 87.6\% to 24.4\%. To understand this phenomenon, we analyze the value-output vectors in these heads and discover that the vectors at each label position contain substantial information about the corresponding labels. Furthermore, we observe that the prediction shift from "foo" to "bar" is due to the respective reduction and increase in these heads' attention scores at "foo" and "bar" positions. Therefore, we propose a hypothesis for ICL: in in-context heads, the value-output matrices extract label features, while the query-key matrices compute the similarity between the features at the last position and those at each label position. The query and key matrices can be considered as two towers that learn the similarity metric between the last position's features and each demonstration at label positions. Using this hypothesis, we explain the majority label bias and recency bias in ICL and propose two methods to reduce these biases by 22\% and 17\%, respectively.
\end{abstract}

\section{Introduction}
In-context learning (ICL) is an emergent ability \cite{wei2022emergent} of large language models \cite{brown2020language,ouyang2022training,touvron2023llama}. By using some demonstration-label pairs as prompts, ICL performs well without updating parameters on many tasks, such as machine translation \cite{sia2023context}, complexity reasoning \cite{li2023towards}, compositional generalization \cite{zhou2022least} and information extraction \cite{he2023icl}.

Because the mechanism of ICL remains unclear, many studies focus on understanding how ICL works. \citet{pan2023context} find that ICL can be disentangled into task recognition (TR) and task learning (TL). TR does not rely on the demonstration-label mappings because the roles of demonstrations and labels are helping the model know "what is the task". In this situation, the model have similar predictions when the mappings are wrong \cite{min2022rethinking}, because the predictions are based on pre-trained priors. On the other hand, TL relies on the demonstration-label mappings because the semantic priors are removed. For example, in an ICL sentiment classification task, if the labels are "positive/negative", the task is TR. If the labels are "foo/bar", the task is TL because the labels are semantically-unrelated \cite{wei2023larger}. \citet{wang2023label} analyze the information flow by averaging all attention heads and find the label words are anchors to merge the semantic information of corresponding demonstrations in shallow layers, and information is extracted from label words to the final prediction in deep layers.

\begin{figure}[htb]
\vspace{-10pt}
\begin{center}
\centerline{\includegraphics[width=\columnwidth]{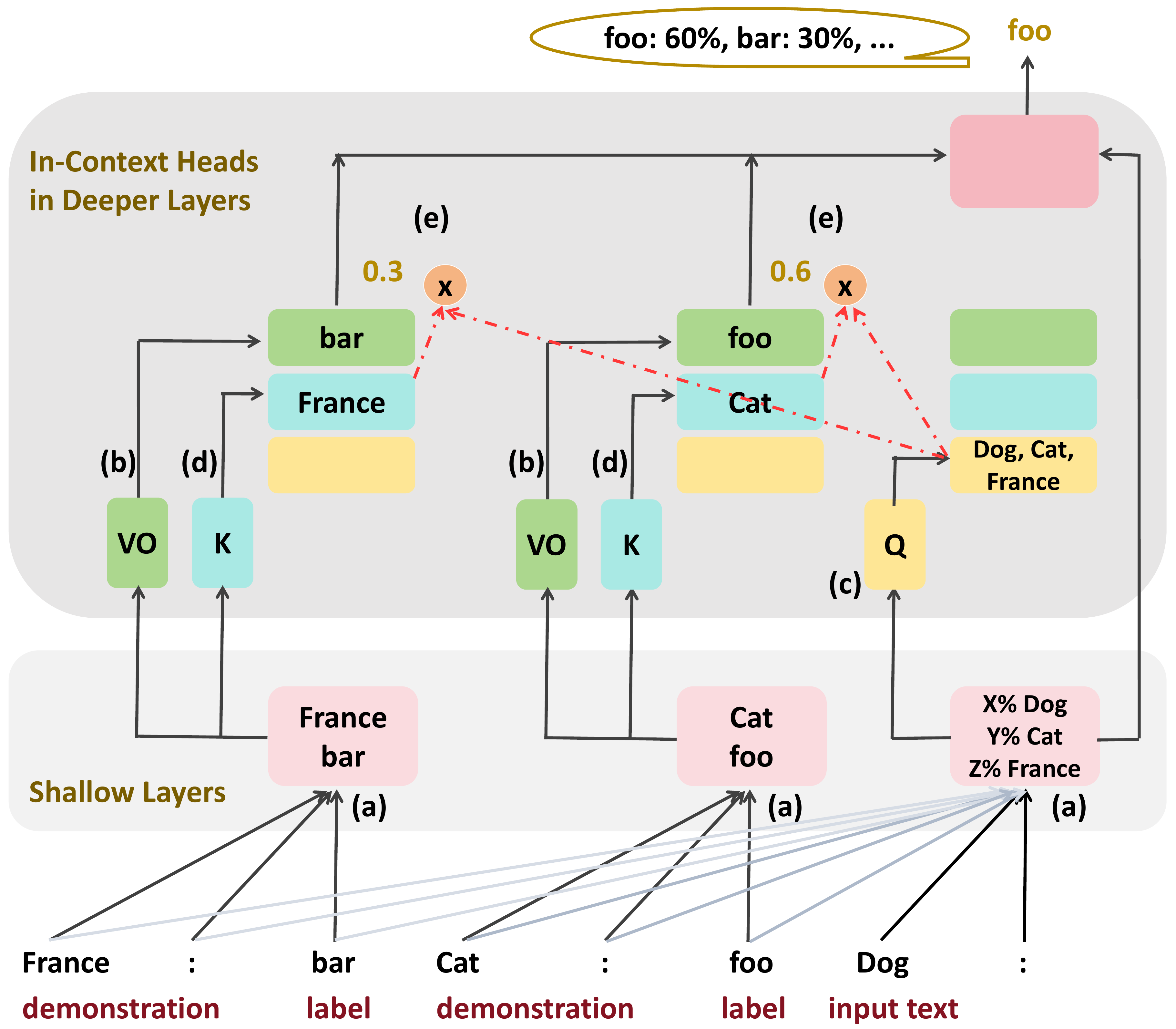}}
\caption{Hypothesis of ICL mechanism. (a) Shallow layers merge features into label positions and last position. In in-context heads, (b) value-output matrix VO extracts label information. (c) Query matrix Q and (d) key matrix K compute the (e) similarity scores between last position and each demonstration, deciding how much label information is transferred into the last token.}
\label{figure1}
\end{center}
\vspace{-10pt}
\end{figure}

Although these studies are important for understanding ICL, the exact mechanism of ICL remains a mystery for several reasons. Firstly, the information flow is typically observed as an average across each head, but understanding ICL requires exploring the precise importance of each head. Secondly, each head has a query matrix, key matrix, value matrix, and output matrix; it is essential to study the role of each matrix in detail. Lastly, ICL is plagued by issues such as majority label bias and recency bias, and how to explain and mitigate these biases has not yet been thoroughly investigated.

In this paper, we address these issues by identifying important heads for ICL and studying the roles of each matrix within these heads. Using two methods, we identify 12 important heads (named in-context heads) that significantly affect ICL accuracy across five datasets, reducing it from 87.6\% to 24.4\% on average. Intervening in 6 heads (fooheads) decreases the probabilities of "foo", while intervening in the other 6 heads (barheads) reduces the probabilities of "bar". To explore the reason of this phenomenon, we study these heads' value-output vectors computing by value-output matrices, and find that the vectors on label positions contain much information about the corresponding labels. Moreover, we observe the attention scores in the in-context heads when predictions shift from "foo" to "bar", and find that the attention scores at "foo" positions decrease, while the attention scores at "bar" positions increase. Based on these observations, we propose a hypothesis for ICL, as shown in Figure 1: in in-context heads, value-output matrices extract label information ("foo"/"bar") from corresponding labels, and query-key matrices compute the similarity between the last position and each label position. The query and key matrices can be regarded as two towers for learning the similarity between the features at last position and each demonstration at label positions. The greater the similarity, the higher the probability of the corresponding label.

Based on this hypothesis, we explore the reason why ICL has majority label bias \cite{zhao2021calibrate} and recency bias \cite{lu2021fantastically}. The existing of majority label bias matches our hypothesis: query and key matrices compute the attention weights between the last position and each demonstration, so the sum of one label's attention weights is larger when this label is related to more demonstrations. About recency bias, we hypothesize that it is caused by the influence of positional embedding during attention score computation in both shallow and deep layers. Based on our analysis, we propose two methods for reducing these biases. For majority label bias, we increase the attention weight of the imbalanced label's position in in-context heads, and the majority label bias reduces 22\%. For recency bias, we remove the affect of position embedding in in-context heads, and the recency bias reduces 17\%. Our code and data will be released on \url{https://github.com/zepingyu0512/in-context-mechanism}.

\section{Related Work}
\subsection{Understanding ICL}
Many studies have explored the mystery of ICL. \citet{min2022rethinking} find that randomly replacing the ground truth labels does not hurt performance much. \citet{wei2023larger} argue the reason of this phenomenon is the model can rely on semantic priors. Therefore, they study semantically-unrelated label ICL by transferring the labels into "foo" and "bar" and find that the performance is related to the demonstration-label mapping. \citet{pan2023context}, disentangle ICL into task recognition (TR) and task learning (TL) to explain this phenomenon. \citet{chan2022data} demonstrate that the ICL ability is obtained when training data have enough rare classes. \citet{liu2021makes} argue that selecting the closest neighbors as demonstrations can enhance ICL ability. \citet{gonen2022demystifying} propose choose low perplexity demonstrations to increase the performance of ICL. \citet{dong2022survey} conclude these methods in a survey for ICL. \citet{wang2023label} find the label words are anchors to extract demonstrations in shallow layers, and the last position extracts information from label words in deep layers.

Some studies try to explain ICL theoretically. \citet{xie2021explanation} argue that ICL ability is gained when the pretraining distribution is a mixture of HMMs, and they explain ICL as implicit Bayesian inference. \citet{garg2022can} prove that transformers can learn linear functions by ICL. \citet{akyurek2022learning} find transformers can learn linear regression functions and hypothesize that ICL can implement standard learning algorithms implicitly. \citet{li2023closeness} explore the softmax regression and find that attention-only transformers are similar with gradient descent models. \citet{von2023transformers} and \citet{dai2022can} regard ICL as meta-learning and argue that ICL does gradient descent implicitly.

\subsection{Mechanistic Interpretability}
The goal of mechanistic interpretility \cite{Chris2022,nanda2023progress} is to reverse engineer the circuits from inputs to outputs. One common method is to apply gradient-based methods \cite{sundararajan2017axiomatic,kindermans2019reliability} or causal tracing methods \cite{pearl2001direct,vig2020investigating,meng2022locating} to analyze the importance of different attention heads and hidden states. \citet{olsson2022context} find that induction heads in attention layers are helpful for copying words from the input sequence (e.g. [X][Y]...[X] -> [Y]). \citet{wang2022interpretability} interpret the circuits on indirect object identification task in GPT2. \citet{hanna2023does} studies how GPT2 computes greater-than by constructing a computational graph of head node and MLP node.

Another common method for mechanistic interpretability is the logit lens \cite{nostalgebraist2020}, whose idea is to analyze the hidden vectors in unembedding space (also named vocabulary space). Many studies have found that the parameters in transformers are interpretable when projecting into vocabulary space \cite{elhage2021mathematical,geva2022transformer,dar2022analyzing,yu2024interpreting}.

\section{Hypothesis for ICL Mechanism}
Our hypothesis is motivated by a case study in Section 3.1. We find that ICL performance can be affected much by only 1\% heads, where some can enhance the probabilities for "foo" and others for "bar" (Section 3.2). To understand why this happens, we analyze the value-output vectors and attention scores in Section 3.3 and find that value-output matrices extract the label information and attention scores computed by query-key matrices control the label information flow. At last, we discuss our hypothesis for ICL in Section 3.4.

\subsection{Hypothesis Motivated by Case Study}
Our hypothesis and analysis is motivated by a case study in GPT2-large \cite{radford2019language}. We design a simple ICL case for word classification: \textbf{"love : bar like : bar eight : foo two : foo one :"}, where the model's prediction is \textbf{"foo"}. In this case, "foo" is the semantic-unrelated label for "number" and "bar" is for "sentiment". We propose a locate-and-project method for case study: we first locate the most important heads using the method discussed in Section 3.2, then project the vectors on label and last positions into vocabulary space by multiplying each vector $v$ and the unembedding matrix $E_u$, following \citet{dar2022analyzing}:
\begin{equation}
    D_v = softmax(E_u \, v)
\end{equation}
Top tokens of value-output vectors and key vectors at label positions and query vector at last position in $a_{22}^0$ (layer22, head0) are shown in Table 1.

\begin{table}[htb]
\centering
\begin{small}
\begin{tabular}{lp{5.5cm}}
\toprule
\textbf{position} & \textbf{top words in vocabulary space} \\
\midrule
2-value & \textbf{BAR}, Barron, Barrett, Band, Bray, \textbf{Bars}, Baron, \textbf{Bar}, Bay, Boyd\\
5-value & \textbf{BAR}, Barron, Barrett, Baron, \textbf{Bar}, Band, Barbie, Barbar, Bard\\
8-value & \textbf{foo}, \textbf{Foo}, FO, fo, Foley, Fresno, FDR, fascists\\
11-value & \textbf{foo}, \textbf{Foo}, fo, FO, fascists, FDR, Foley, Goo, fascists\\
2-key & \textbf{kisses}, \textbf{goddess}, \textbf{love}, \textbf{charms}, idol, stress, nobles, \textbf{happiness}\\
5-key & style, oriented, +++, \textbf{like}, indo, height, Lover, xual, dont, foo \\
8-key & foo, mc, blah, happ, avg, french, omega, prod, english, google, height, neigh \\
11-key & foo, mc, infinity, omega, \textbf{three}, \textbf{two}, repeat, \textbf{twelve}, 666, \textbf{Three}, \textbf{thirds}, \textbf{five}, \textbf{sixteen} \\
13-query & \textbf{first}, end, only, no, all, given, person, certain, call, same, short, long, \textbf{1}, \textbf{one}, value \\
\bottomrule
\end{tabular}
\end{small}
\caption{Top tokens at label positions and last position.}
\vspace{-10pt}
\end{table}

Label positions' value-output vectors contain concepts about the labels, and their key vectors contain the corresponding demonstrations. For example, the label at position 2 is "bar" and the value-output vector contains "BAR, Bars, Bar". Its key vector's top tokens are related to the corresponding demonstration "love". The last position have concepts about the input text "one". Hence, we hypothesize that value-output matrices extract the label information and query-key matrices compute the similarity between the last position (encodes the input text) and each label position (encodes demonstration). We also note interpretable results in sentence classification cases, detailed in Appendix A.

\subsection{Identifying Important Heads for ICL}
\paragraph{Datasets and models.} We conduct the experiments on five sentence classification datasets, including financial phrasebank (Financ) \cite{Malo2014GoodDO}, AG’s news topic classification (AGnews) \cite{zhang2015character}, Amazon reviews (Amazon) \cite{mcauley2013hidden}, Hate Speech Detection (ETHOS) \cite{mollas2020ethos}, and Stanford Sentiment Treebank binary (SST2) \cite{socher2013recursive}. We conduct experiments on Llama-7B \cite{touvron2023llama} with 32 layers (32 heads per layer), and GPT-J \cite{gpt-j} with 28 layers (16 heads per layer).

\begin{table}[htb]
\centering
\begin{small}
\begin{tabular}{cccccc}
\toprule
 & Financ & AGnews & Amazon & ETHOS & SST2 \\
\midrule
foo & 90.6 & 96.6 & 84.2 & 69.0 & 89.2 \\
bar & 99.8 & 100.0 & 85.6 & 73.2 & 88.8 \\
\midrule
foo & 97.6 & 99.6 & 65.2 & 54.2 & 90.4 \\
bar & 98.6 & 83.2 & 98.8 & 92.8 & 97.2 \\
\bottomrule
\end{tabular}
\end{small}
\caption{ICL accuracy (\%) with correct label "foo"/"bar" in Llama (first block) and GPT-J (second block).}
\vspace{-10pt}
\end{table}

Inspired by \citet{pan2023context} and \citet{wei2023larger} that task learning is the emergent ability of large language models (LLMs), we replace the labels with semantic-unrelated labels "foo" and "bar" to study the mechanism of ICL task learning ability. In each dataset, we randomly sample two sentences with each label, and propose the ICL sentence: \textbf{"S0 : bar S1 : bar S2 : foo S3 : foo S4 :"} with correct label "foo" and \textbf{"S0 : foo S1 : foo S2 : bar S3 : bar S4 :"} with correct label "bar", where S0 and S1 have the same label, and S2, S3, S4 have the other label. We randomly sample 1,000 sentences in each dataset. The accuracy when correct labels are "foo" and "bar" are shown in Table 2, which indicate that the ICL ability exists in most datasets. 

\paragraph{Methods.} We apply two methods to identify the important heads for ICL. Firstly, we use causal tracing methods \cite{pearl2001direct,vig2020investigating} and intervene each head in deep layers by setting the head's parameters to zero, and re-calculate the decrease in each dataset. Secondly, following \citet{yu2024locating}, we compute the log probability increase $S_l^h$ of each head to find which heads directly contribute to the final predictions: 
\begin{equation}
S_l^h = log(p(b|o_l^h+Lin_l))-log(p(b|Lin_l))
\end{equation}
where $b$ is the predicted label ("foo"/"bar"), $Lin_l$ is $lth$ layer's input, and $o_l^h$ is the head output vector on layer $l$, head $h$. The probability is calculated by multiplying the vector with the unembedding matrix $E_u$ (Eq.1). If the score is large, the head is useful for increasing the probability of label $b$. We identify the heads rank top10 in both methods, and there are 6 important "fooheads" affecting "foo" and 6 important "barheads" affecting "bar" in both model. The average accuracy change when intervening the fooheads and barheads is shown in Table 3. When intervening the fooheads, datasets with correct label "foo" show a significant decrease in accuracy, while those with correct label "bar" experience a substantial increase in accuracy. When masking in the barheads, datasets with correct label "bar" show a significant decrease in accuracy, while those with correct label "foo" experience a substantial increase in accuracy. Therefore, our identified fooheads and barheads are important for predicting "foo" and "bar", respectively. We name these heads "in-context heads".

\begin{table}[htb]
\centering
\begin{small}
\begin{tabular}{ccccc}
\toprule
\multicolumn{1}{c}{} & \multicolumn{2}{c}{correct label : foo} & \multicolumn{2}{c}{correct label : bar}\\ 
 & fooheads & barheads & fooheads & barheads \\
\midrule
Llama & 86.0/0.01 & 86.0/99.3 & 89.0/99.2 & 89.0/35.4 \\
GPT-J & 81.4/10.6 & 81.4/98.9 & 94.1/99.9 & 94.1/51.4 \\
\bottomrule
\end{tabular}
\end{small}
\caption{Origin/intervened accuracy (\%) when intervening fooheads and barheads in Llama and GPT-J.}
\vspace{-10pt}
\end{table}

\subsection{Analyzing Value-Output Vectors and Attention Scores in In-context Heads}
Head output $o$ in Eq.1 is computed by the weighted sum of value-output vectors $vo$ on all positions $p$: 
\begin{equation}
    o = \sum_{p=0}^{T-1} \alpha^p \cdot vo^p
\end{equation}
where $T$ is the length of the input text. $\alpha$ is the attention score computed by the softmax function on the inner product of last position's query vector and each position's key vector. $vo$ is computed by the linear transform of value-output matrices on each position's layer input. To explore the importance of label positions in each in-context head, we investigate sentences with correct label "foo", and compute the logit minus score $M$ at "foo" and "bar" positions' weighted value-output vectors:
\begin{equation}
M = log(p(foo|\alpha^{p} \cdot vo^{p})) - log(p(bar|\alpha^{p} \cdot vo^{p})) 
\end{equation}
If $M$ is larger than zero, the vectors are important for enhancing "foo" probability. On the contrary, they are important for enhancing "bar" probability. The average logit minus scores at "foo" positions (fp) and "bar" positions (bp) in fooheads (fh) and barheads (bh) are shown in Table 4. In both models, foo positions contain much information about "foo" in fooheads, and bar positions contain much information about "bar" in barheads. Furthermore, the proportion between label positions' logit minus scores and the in-context heads' logit minus scores is 99.1\%. Therefore, the reason fooheads/barheads affect probabilities of "foo"/"bar" is due to the information saved at "foo"/"bar" positions' weighted value-output vectors $\alpha \cdot vo$. 

\begin{table}[htb]
\centering
\begin{small}
\begin{tabular}{cccccc}
\toprule
 & Financ & AGnew & Amaz & ETHOS & SST2 \\
\midrule
fh-fp & 0.29 & 0.32 & 0.30 & 0.30 & 0.32 \\
fh-bp & -0.02 & -0.05 & -0.04 & -0.04 & -0.04 \\
bh-fp & -0.05 & -0.03 & -0.03 & -0.02 & -0.04 \\
bh-bp & -0.11 & -0.08 & -0.14 & -0.14 & -0.12 \\
\midrule
fh-fp & 0.26 & 0.23 & 0.26 & 0.28 & 0.31 \\
fh-bp & 0.00 & -0.01 & 0.00 & -0.01 & -0.01 \\
bh-fp & -0.07 & -0.05 & -0.06 & -0.06 & -0.06 \\
bh-bp & -0.16 & -0.17 & -0.20 & -0.23 & -0.18 \\
\bottomrule
\end{tabular}
\end{small}
\caption{Logit minus of weighted value-output vectors at "foo"/"bar" positions (fp, bp) in fooheads/barheads (fh, bh) in Llama (first block) and GPT-J (second block).}
\vspace{-10pt}
\end{table}

To explore the roles of query-key matrices and value-output matrices, we compute the attention scores and the value-output vectors' logit minus scores (removing $\alpha^p$ in Eq.4). The average scores across all datasets are shown in Table 5.

\begin{table}[htb]
\centering
\begin{small}
\begin{tabular}{ccccc}
\toprule
\multicolumn{1}{c}{} & \multicolumn{2}{c}{fooheads} & \multicolumn{2}{c}{barheads}\\ 
 & foopos & barpos & foopos & barpos \\
\midrule
attn & 0.742 & 0.047 & 0.369 & 0.195 \\
minus & 0.613 & -0.574 & -0.075 & -0.658 \\
\midrule
attn & 0.540 & 0.037 & 0.219 & 0.203 \\
minus & 0.958 & 0.099 & -0.253 & -1.656 \\
\bottomrule
\end{tabular}
\end{small}
\caption{Attention score and logit minus at "foo"/"bar" positions in fooheads/barheads in Llama (first block) and GPT-J (second block), averaged on all datasets.}
\vspace{-10pt}
\end{table}

Both query-key matrices and value-output matrices can affect the probabilities. In Llama fooheads, the query-key matrices play large roles for predicting "foo". The value-output matrices can extract both "foo->foo" and "bar->bar", since the absolute values of logit minus scores at "foo" and "bar" positions are similar. In GPT-J fooheads, both query-key matrices and value-output matrices play large roles for enhancing "foo". In Llama barheads and GPT-J barheads, value-output matrices play larger role than query-key matrices for predicting "bar".

To explore how the predictions change from "foo" to "bar", we compare the sentences \textbf{"S0 : bar S1 : bar S2 : foo S3 : foo S4 :"} and \textbf{"S0 : foo S1 : foo S2 : bar S3 : bar S4 :"} in each dataset. We compute the change of absolute value on weighted value-output vectors' logit minus scores (minus-w), value-output vectors' logit minus scores (minus), and attention scores, shown in Table 6.

\begin{table}[htb]
\centering
\begin{small}
\begin{tabular}{ccccc}
\toprule
\multicolumn{1}{c}{} & \multicolumn{2}{c}{fooheads} & \multicolumn{2}{c}{barheads}\\ 
 & foopos & barpos & foopos & barpos \\
\midrule
minus-w & -12.1\% & +54.4\% & -47.9\% & +47.2\% \\
minus & +26.5\% & -21.4\% & +41.8\% & -10.5\% \\
attn & -21.8\% & +124.8\% & -44.4\% & +91.4\% \\
\midrule
minus-w & -40.4\% & +408.5\% & -51.7\% & +55.1\% \\
minus & +13.9\% & +32.0\% & +44.2\% & -17.6\% \\
attn & -43.0\% & +237.8\% & -46.1\% & +86.0\% \\
\bottomrule
\end{tabular}
\end{small}
\caption{Change of attention score and logit minus at "foo"/"bar" positions in fooheads/barheads in Llama (first block) and GPT-J (second block) on all datasets.}
\vspace{-10pt}
\end{table}

The prediction shift is caused by the change of weighted value-output vectors' logit minus scores. When changing the labels, fooheads' foo positions contain less information about "foo", and barheads' bar positions contain more information about "bar". The "foo" decrease at fooheads' "foo" positions and the "bar" increase at barheads' "bar" positions cause the probability change from "foo" to "bar".

The attention scores change significantly when the predictions shift from "foo" to "bar". Attention scores at fooheads' "foo" positions decrease substantially, while those at barheads' "bar" positions increase markedly. Comparatively, the change direction of the value-output vectors' logit minus scores does not show a relevant trend with the logit minus scores of the weighted value-output vectors. Therefore, we hypothesize that the change of attention scores within in-context heads is the primary cause for the prediction shift from "foo" to "bar".

\subsection{Proposed Hypothesis and Discussion}
For better understanding, we list the evidence of existing studies and previous sections: a) \citet{wang2023label} demonstrate that the label positions ("foo", "bar") extract corresponding demonstrations' features in shallow layers. b) In Section 3.2, we find that in deep layers there are a few fooheads important for predicting "foo" and barheads for "bar". c) Table 4 proves that the "foo" positions in fooheads and the "bar" positions in barheads contain much information for predicting "foo" and "bar", respectively. d) The experiments in Table 5 demonstrate that both query-key matrices and value-output matrices can affect the information storage. e) Table 6's results prove that the change of attention scores within fooheads and barheads is the primary cause for the prediction shift from "foo" to "bar"  when reversing the demonstrations' labels.

Based on these findings, we conclude our hypothesis: In shallow layers, the label positions extract features from the corresponding demonstrations (hypothesized from evidence a), while the last position encodes information of the input text and previous demonstrations/labels (X\% input text + Y\% near demonstrations + Z\% far demonstrations). In deep layers' in-context heads, the value-output matrices extract the label features into value-output vectors (hypothesized from evidence b and c). For example, fooheads extract "foo->foo" and barheads learn "bar->bar". The query-key matrices compute the similarity between the last position's features and each label position's features. When the labels change from "foo" to "bar", the change of last position features causes the similarity scores change and the prediction shift (hypothesized from evidence d and e). For instance, the fooheads' similarity scores at foo positions change from SIM((X+Y)\%foo, foo) to SIM(Z\%foo, foo), and the barheads' similarity scores at bar positions change from SIM(Z\%bar, bar) into ((X+Y)\%bar, bar). Hence, the foo positions' attention scores decrease in fooheads and the bar positions' attention scores increase in barheads, causing the probability change from "foo" to "bar". 

If considering all the in-context heads together, the overall value-output matrices can learn both "foo->foo" and "bar->bar". Under our hypothesis, the query and key matrices can be regarded as two towers computing the semantic similarity between the last position's features and each label position's demonstration features. If the similarity score is large, more corresponding label information is incorporated, enhancing the probability of that label. There are four modules related to the ICL ability. 

\textbf{a) Information extraction ability of shallow layers.} Shallow layers can be regarded as feature extraction modules. The ability of extracting corresponding demonstrations and the input text decides the quality of features. 

\textbf{b) Value projection ability of in-context heads' value-output matrices.} If the value projection ability is good enough, the in-context heads should project "foo" and "bar" together and fairly.

\textbf{c) Metric learning ability of in-context heads' query and key matrices.} The query and key matrices might be the most important module, because they should learn computing different metrics using the same matrices. If different ICL tasks share the same in-context heads, the query and key matrices should learn these metrics jointly.

\textbf{d) Numbers and parameters of in-context heads.} If we regard one in-context head as a two-tower model for metric learning, the parameters of the head are directly related to the learning ability. At the same time, different in-context heads can be regarded as voting or ensemble models, so the head number also controls the learning ability. 

\section{Understanding Majority Label Bias and Recency Bias in ICL}
There are several phenomena of ICL that haven't been explained. \citet{zhao2021calibrate} demonstrate that models tend to predict majority labels and the labels near the input text. \citet{lu2021fantastically} also find that changing the demonstration order can affect predictions a lot. Based on our hypothesis, we explore why ICL has majority label bias (in Section 4.1) and recency bias (in Section 4.2).

\subsection{Understanding Majority Label Bias}
According to our hypothesis, it is reasonable that the model tends to predict majority labels, because the label information flow is controlled by the similarity between last position and each label position. When a label has high frequency, the sum of similarity scores will be larger, thus the probability of this label is larger in final prediction. We design an imbalanced dataset to verify this. For each sentence with correct label "foo", we remove the last demonstration and label. For example, \textbf{"S0 : bar S1 : bar S2 : foo S3 : foo S4 :"} is changed into \textbf{"S0 : bar S1 : bar S2 : foo S4 :"}. We compute the sum of attention weights on "foo" positions in fooheads and "bar" positions in barheads on the imbalanced datasets and the original datasets, averaged on all five datasets. The changing of attention scores at "foo" positions and "bar" positions in both models are shown in Figure 2.

\begin{figure}[htb]
	\centering
	\subfigure{
		\begin{minipage}[b]{0.47\textwidth}\includegraphics[width=0.47\textwidth]{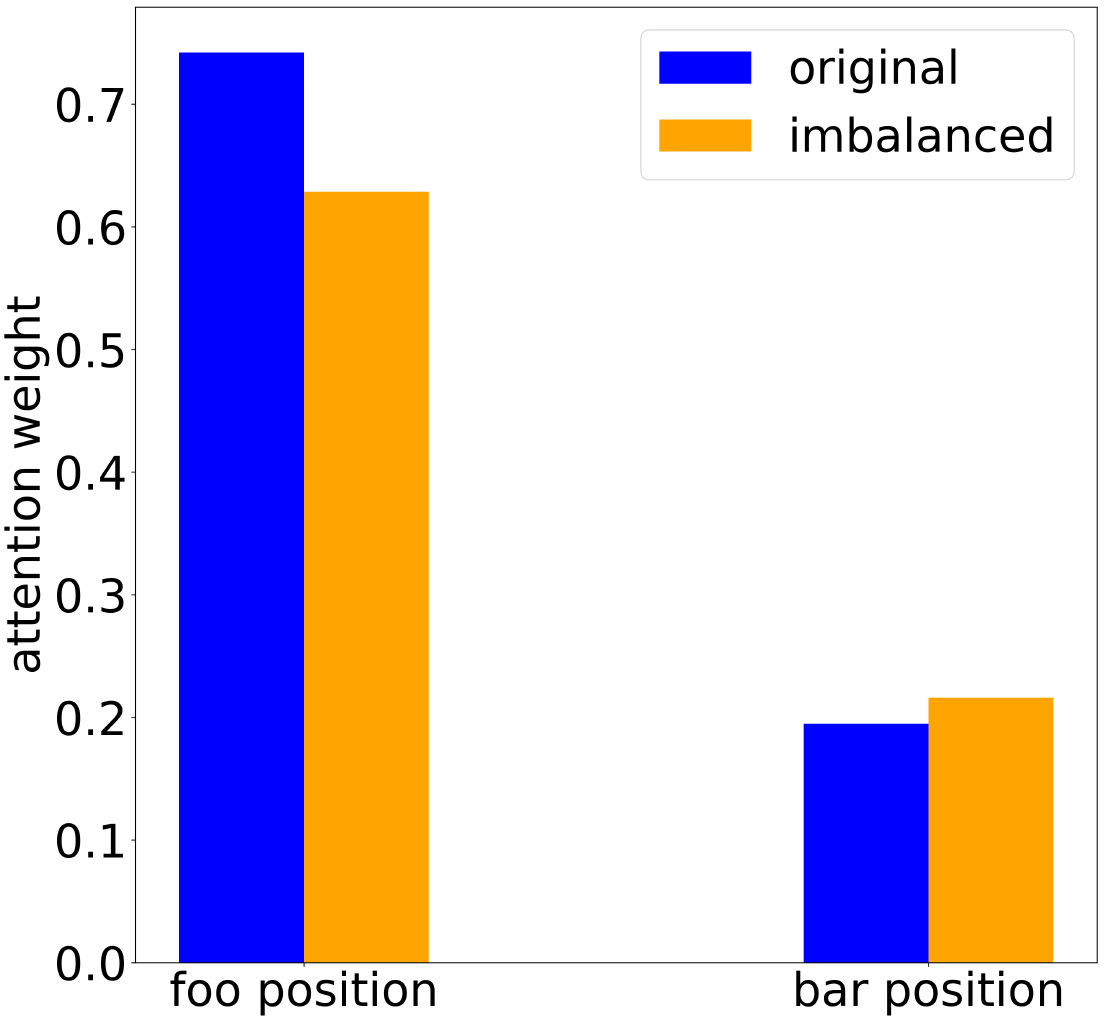}
        \includegraphics[width=0.47\textwidth]{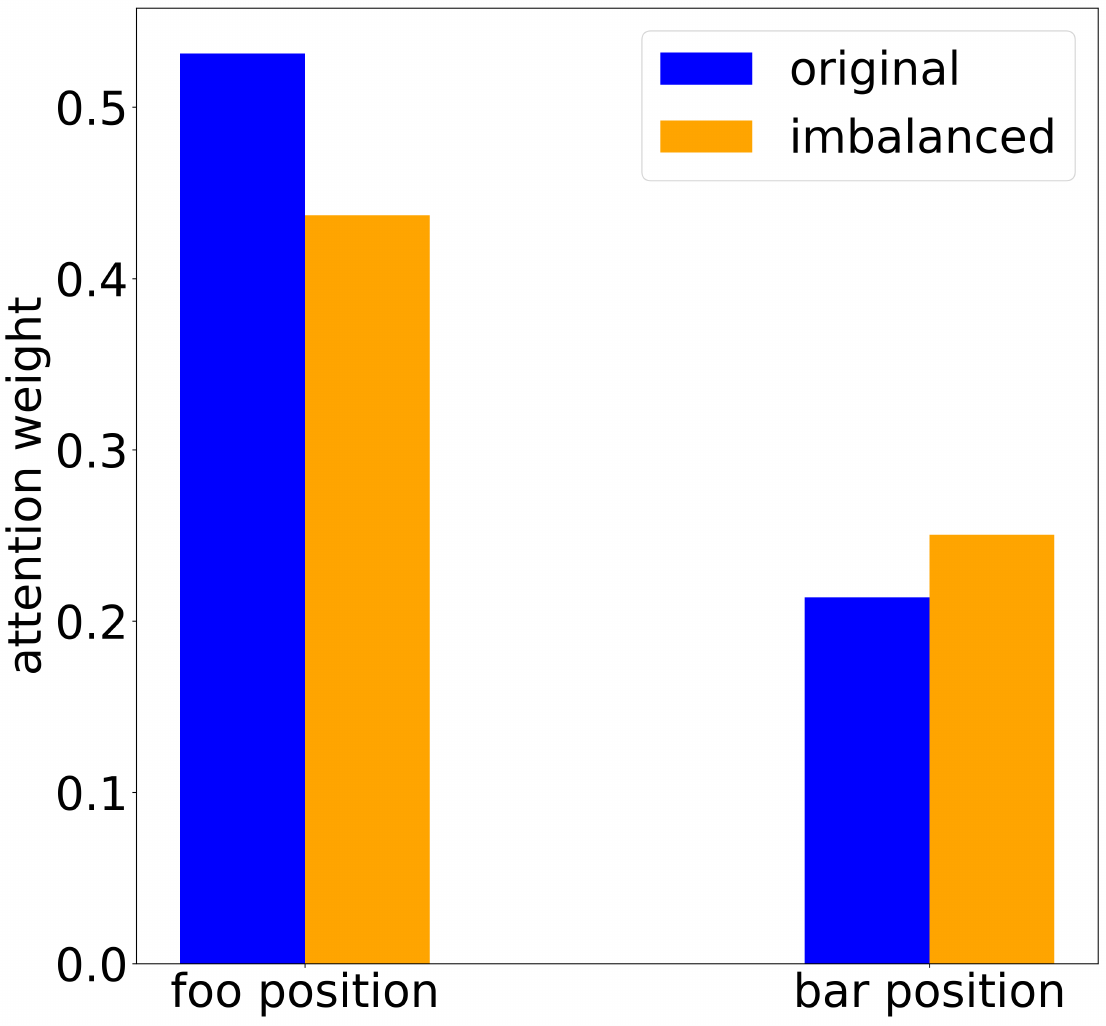}
		\end{minipage}
	}
	\caption{Attention scores on foo positions in fooheads and bar positions in barheads, on original dataset and imbalanced dataset in Llama (left) and GPT-J (right).}
\end{figure}

In both models, the sum of attention weights on "foo" positions decrease on the imbalanced dataset. On the contrary, the attention weights on "bar" positions increase. The results meet our analysis. The attention weights are computed by a softmax function, so when a "foo" demonstration and its label are removed, the sum of attention weights on "foo" positions will decrease, and that on "bar" positions will increase.

\subsection{Understanding Recency Bias}
The ICL performance is extremely sensitive to the demonstration order. We hypothesize that the recency bias is caused by the influence of positional embeddings on the attention score computation in both shallow layers and deep layers. The attention score is calculated by applying a softmax function to the product of the last position's query vector and each label position's key vector. These query and key vectors are derived from the layer input, which is a combination of the positional embedding, the word embedding, and the output vectors from previous attention layers and feed-forward network (FFN) layers. Therefore, a "position term" consistently influences the attention scores.

The feature extraction of last position is related to the attention scores in shallow layers' heads. Due to the influence of positional embedding, the model tends to extract varying amounts of features at different positions. Let us consider the case \textbf{"S0 : bar S1 : bar S2 : foo S3 : foo S4 :"}. The last position contains X\% S4 + Y\% (S2+S3) and Z\% (S0+S1), simplified into (X+Y)\% foo + Z\% bar. If the demonstration order is changed into \textbf{"S2 : foo S3 : foo S0 : bar S1 : bar S4 :"}, the last position will contain X\% S4 + Z\% (S0+S1) + Y\% (S2+S3), simplified into (X+Z)\% foo + Y\% bar. Hence, the final prediction probability will be different between these two sentences if Y and Z are different. If Y is larger than Z, the last position will contain less "foo". Similarly, the influence of positional embeddings also exists in deep layers' heads, which tends to enlarge the attention scores on later positions in these heads. 

We design a reverse dataset to evaluate the difference among different positions. For each sentence \textbf{S0 : bar S1 : bar S2 : foo S3 : foo S4 :}, we transfer it into a reverse sentence \textbf{S2 : foo S3 : foo S0 : bar S1 : bar S4 :}. We compute the average attention score change at "foo" positions in fooheads and "bar" positions in barheads, between the original and the reverse dataset, shown in Figure 3. Moreover, we remove the impact of positional embedding in each in-context head and re-compute the attention scores (original modify and reverse modify in Figure 3).

\begin{figure}[htb]
	\centering
	\subfigure{
		\begin{minipage}[b]{0.47\textwidth}\includegraphics[width=0.47\textwidth]{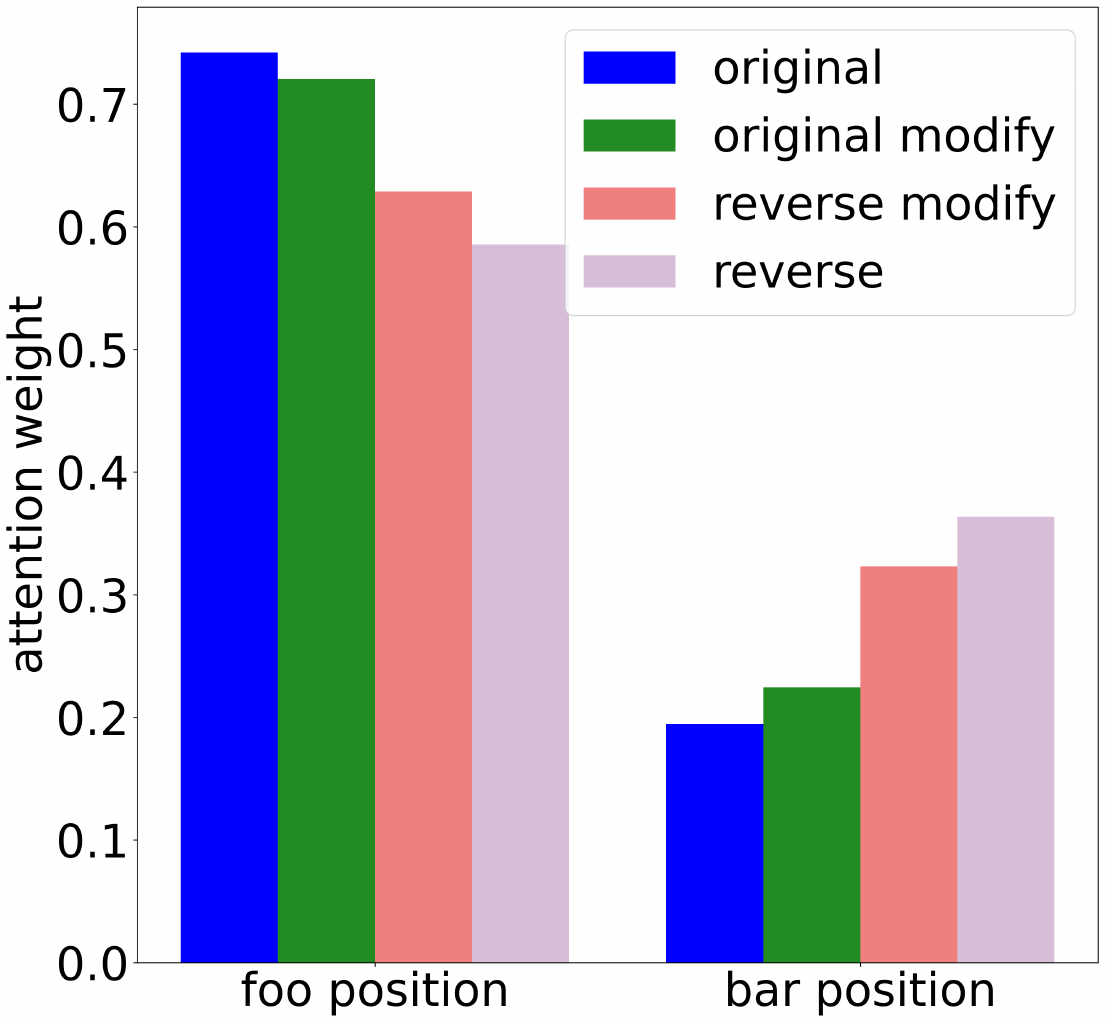}
        \includegraphics[width=0.47\textwidth]{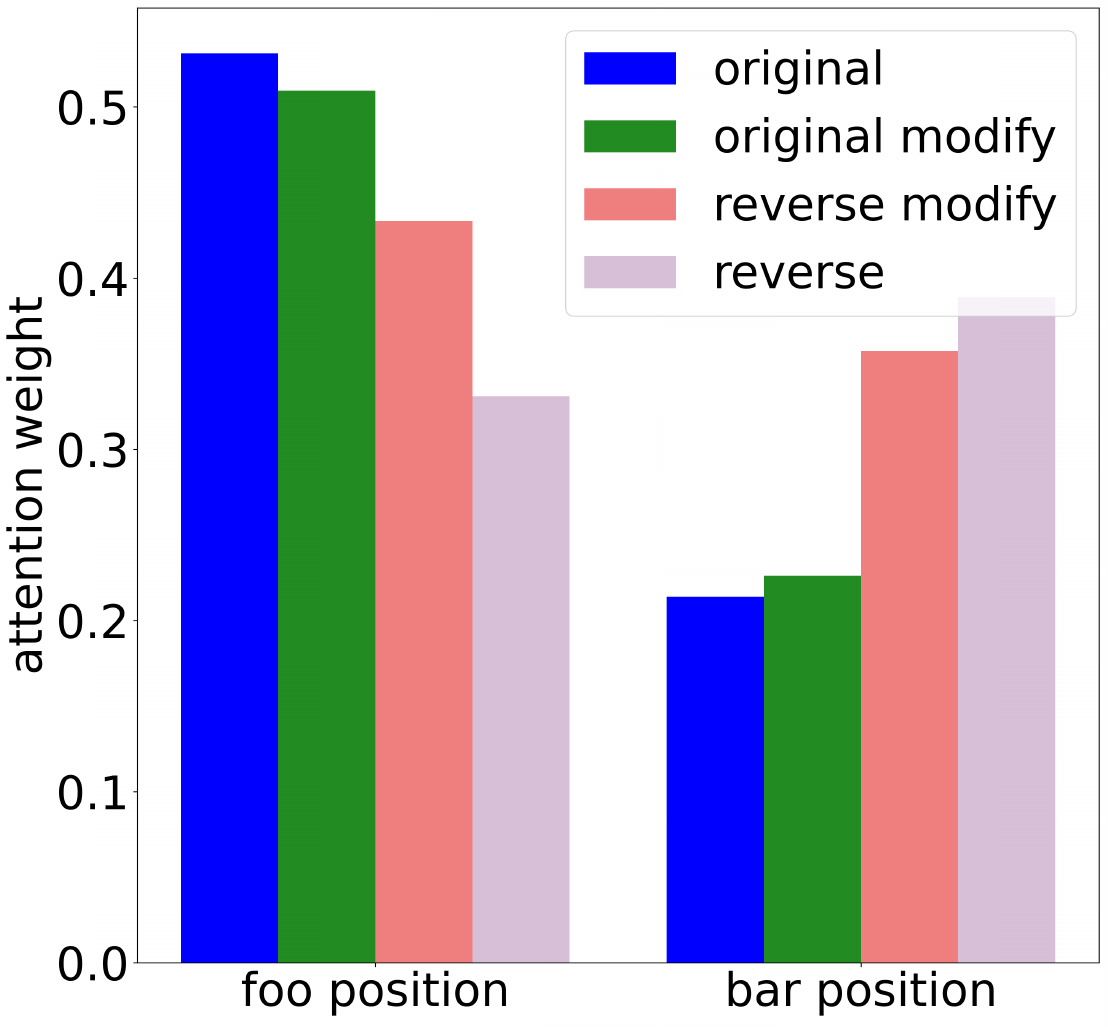}
		\end{minipage}
	}
	\caption{Attention scores on foo positions in fooheads and bar positions in barheads, on original dataset and reverse dataset in Llama (left) and GPT-J (right).}
\end{figure}

Compared with the original dataset, "foo" positions' attention weights decrease and "bar" positions' attention weights increase in the reverse dataset in both models. This result aligns with the observations in previous studies \cite{zhao2021calibrate} that the probability is affected much when reversing the demonstration order. When removing the impact of positional embedding in each head, the near positions' attention scores decrease and the far positions' scores increase. Hence, our hypothesis is verified: the positional term in each head enlarges the attention scores on later positions. After removing the positional term in in-context heads, the attention score is still different between the original dataset and the reverse dataset. This difference is caused by the difference in shallow layers' feature extraction stage. 

To provide a clearer perspective, we illustrate the attention score change on "foo" positions in each foohead and "bar" positions in each barhead. The change of imbalanced dataset and reverse dataset in Llama and GPT-J is shown in Figure 5 and 6, where the first 6 columns are "foo" positions' attention scores in fooheads and the last 6 columns are "bar" positions' scores in barheads. Compared with the original dataset, the attention scores decrease on "foo" positions and increase on "bar" positions in imbalanced dataset and reverse dataset.

\begin{figure}[htb]
\begin{center}
\centerline{\includegraphics[width=0.95\columnwidth]{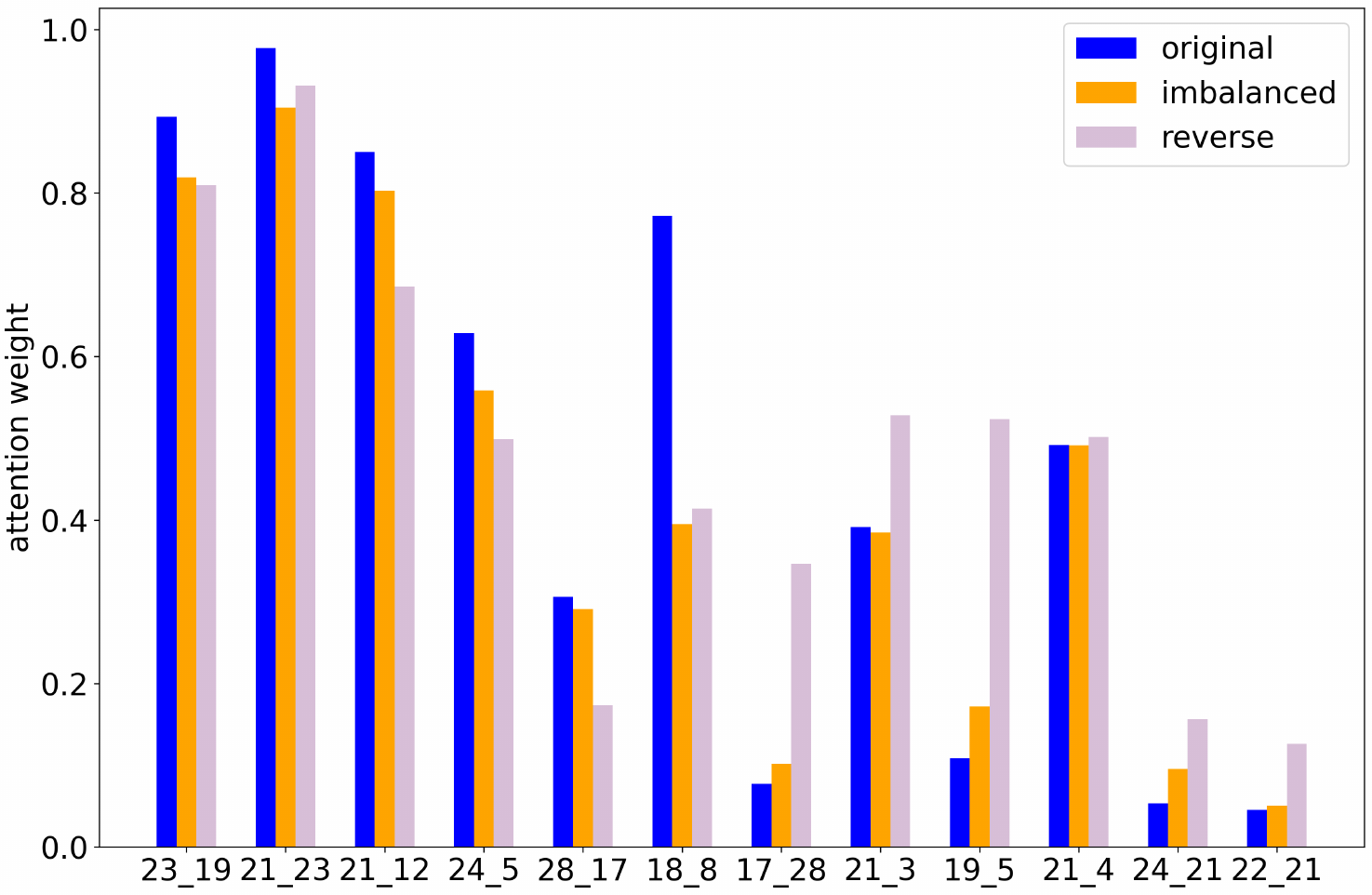}}
\caption{Attention scores on "foo"/"bar" positions in original, imbalanced, and recency datasets in Llama.}
\end{center}
\vspace{-10pt}
\end{figure}

\begin{figure}[htb]
\begin{center}
\centerline{\includegraphics[width=0.95\columnwidth]{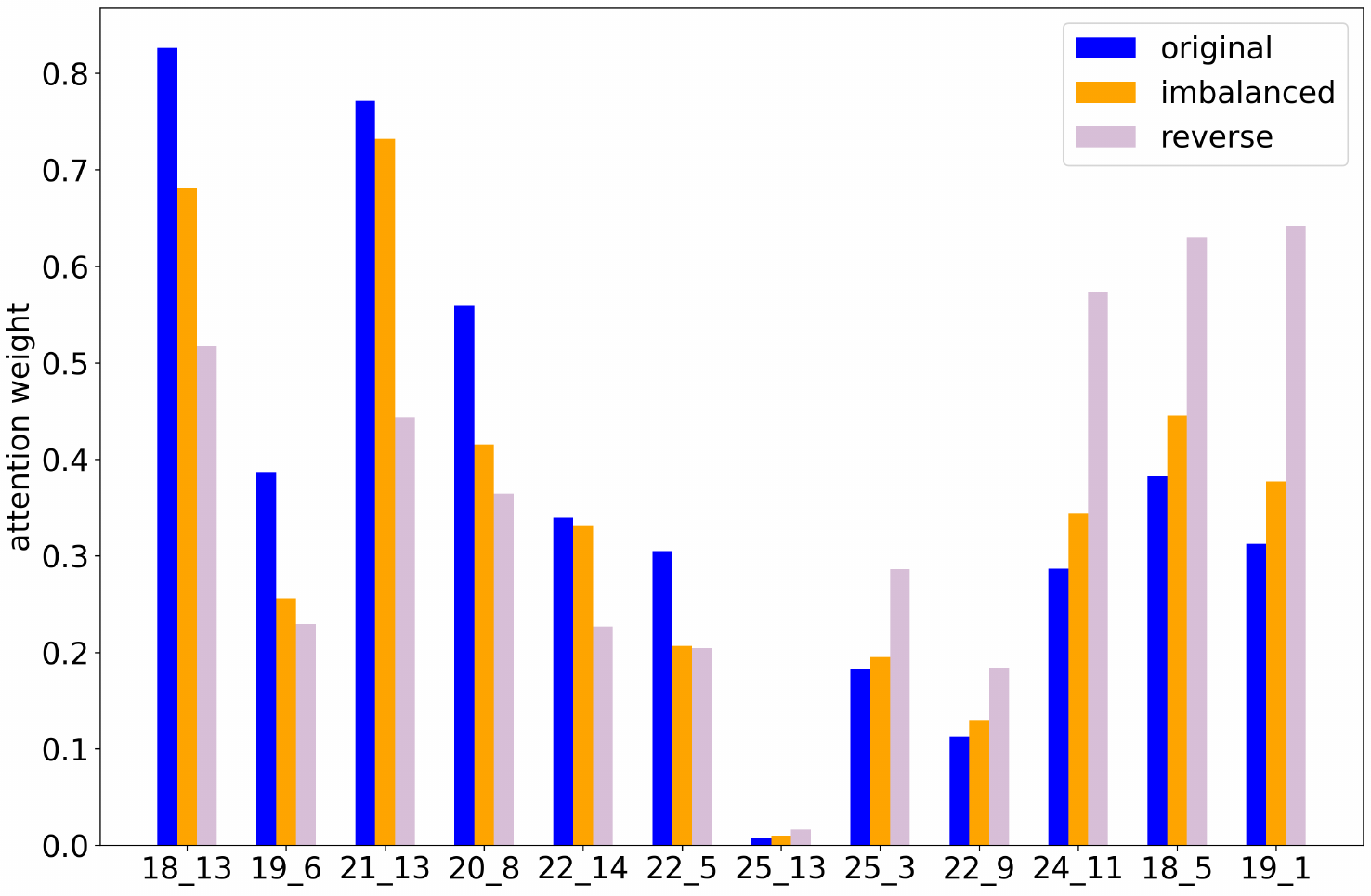}}
\caption{Attention scores on "foo"/"bar" positions in original, imbalanced, and recency datasets in GPT-J.}
\end{center}
\vspace{-10pt}
\end{figure}

\section{Reducing Majority Label Bias and Recency Bias}
In this section, we propose a method for reducing majority label bias in Section 5.1, and propose a method for reducing recency bias in Section 5.2.

\subsection{Reducing Majority Label Bias by Enlarging Imbalanced Label Attention}
According to our analysis in Section 4.1, the majority label bias can be attributed to the lack of attention weights on imbalanced label positions. So we propose a method to reduce the majority label bias by enlarging the imbalanced label positions' attention scores. Specifically, we multiply an amplified score $a$ on the imbalanced label positions' weighted value-output vectors ($a \alpha^p \cdot vo^p$ in Eq.3) and add this vector into the final embedding. $a$ is the product of a constant hyperparameter $a_c$ and a varying score $a_v$, where $a_v$ is the ratio of the larger demonstration number to the smaller demonstration number.

We first make a balanced dataset by randomly sampling 2-4 demonstrations in each label, and randomly set the demonstration order. The correct labels of the balanced sentences are "foo". Then we get a "lackfoo" sentence by randomly removing a "foo" demonstration, and a "lackbar" sentence by randomly removing a "bar" demonstration. Except the results in Financ GPT-J, the accuracy of "lackfoo" dataset is smaller than the balanced dataset due to the lack of "foo" demonstrations, and "lackbar" accuracy is larger than the balanced dataset. 

\begin{table}[htb]
\centering
\begin{small}
\begin{tabular}{cccccc}
\toprule
 & Financ & AGnew & Amaz & ETHOS & SST2 \\
\midrule
before & 0.10 & 0.09 & 0.23 & 0.10 & 0.19 \\
after & 0.07 & 0.05 & 0.17 & 0.07 & 0.15 \\
\midrule
before & 0.04 & 0.03 & 0.05 & 0.05 & 0.08 \\
after & 0.06 & 0.02 & 0.02 & 0.03 & 0.06 \\
\bottomrule
\end{tabular}
\end{small}
\caption{Accuracy change before/after applying our method in Llama (first block) and GPT-J (second block).}
\vspace{-10pt}
\end{table}

Compared to the balanced dataset, we calculate the sum of accuracy change on "lackfoo" and "lackbar" datasets  before and after applying our method with amplified constant score $a_c$ 0.03. The accuracy change is shown in Table 7. On average, the accuracy change reduces 29.1\% in Llama and 14.9\% in GPT-J. The results indicate that our method can reduce the accuracy change caused by the influence of imbalanced demonstrations/labels.

\subsection{Reducing Recency Bias by Removing Positional Embedding Affect}
As discussed in Section 4.2, we find the recency bias is due to the effect of positional embedding on the calculation of attention scores. Hence, in order to reduce the recency bias, we reduce the position term in in-context heads, and re-calculate the output vectors in all in-context heads. This method is similar with adding a shortcut adapter from each in-context head to the final embedding.

\begin{table}[htb]
\centering
\begin{small}
\begin{tabular}{cccccc}
\toprule
 & Financ & AGnew & Amaz & ETHOS & SST2 \\
\midrule
acc-be & 0.37 & 0.42 & 0.26 & 0.22 & 0.30 \\
acc-af & 0.31 & 0.39 & 0.15 & 0.16 & 0.18 \\
attn-be & 0.06 & 0.08 & 0.06 & 0.05 & 0.06 \\
attn-af & 0.03 & 0.06 & 0.04 & 0.03 & 0.03 \\
\midrule
acc-be & 0.39 & 0.27 & 0.45 & 0.41 & 0.40 \\
acc-af & 0.36 & 0.16 & 0.42 & 0.40 & 0.35 \\
attn-be & 0.07 & 0.05 & 0.07 & 0.06 & 0.08 \\
attn-af & 0.04 & 0.03 & 0.05 & 0.04 & 0.05 \\
\bottomrule
\end{tabular}
\end{small}
\caption{Standard deviation of accuracy and attention scores before/after applying our method in Llama (first block) and GPT-J (second block).}
\vspace{-10pt}
\end{table}

We apply this method to the original dataset and three recency datasets with different demonstration orders, detailed in Appendix B. We calculate the standard deviation in accuracy and in-context heads' attention scores before (acc-be, attn-be) and after (acc-af, attn-af) applying our method. The results are shown in Table 8. On average, the accuracy standard deviation reduces 23.4\% in Llama and 10.6\% in GPT-J, and the attention score standard deviation reduces 40.1\% in Llama and 37.7\% in GPT-J. Therefore, removing the positional term in in-context heads is helpful for reducing the recency bias. It is also important to reduce the recency bias during feature extraction in shallow layers, and we leave this exploration in future work.

\section{Conclusion}
We identify the important heads for ICL and analyze the value-output vectors and attention scores in these heads. We propose a hypothesis for the mechanism of ICL. In shallow layers, the demonstrations and input text is captured by the label positions and the last position. In in-context heads, the value-output matrices project the label features into value-output vectors. The query and key matrices can be regarded as two towers learning the similarity between the last position's features and each label position's features. If the similarity score is high, the corresponding label's probability is enlarged. Based on this hypothesis, we interpret why ICL has majority label bias and recency bias. Furthermore, we propose two methods to reduce these biases by 22\% and 17\%. Overall, our study provides a new method and a reasonable hypothesis for understanding the mechanism of in-context learning.

\section{Limitation}
In this paper, we focus on understanding the mechanism in in-context heads in deep layers. It is also important to study how shallow layers transfer features into label positions and the last position. Our hypothesis explains the ICL mechanism for classification tasks. More studies should be done on other ICL tasks, such as chain-of-thought reasoning \cite{wei2022chain}.

Another limitation of our work comes from the attribution method for identifying important heads. Gradient-based methods and causal tracing methods, which calculate a module's impact on the final prediction, are commonly employed for importance attribution. Additionally, many studies utilize saliency score-based methods. In this paper, we apply both causal tracing and saliency score-based methods to identify important heads, and we believe the results in Table 3 support our findings. However, it is important to note that there is no unified method for attributing important modules, and further exploration is needed to design better attribution methods.

\section{Acknowledgements}
This work is supported by the project JPNP20006 from New Energy and Industrial Technology Development Organization (NEDO). This work is supported by the computational shared facility and the studentship from the Department of Computer Science at the University of Manchester. 

\bibliography{anthology,custom}
\bibliographystyle{acl_natbib}
\clearpage
\appendix
\section{Case Study on Sentence Classification}
We analyze a sentence classification case sampled in AGNews dataset. The top tokens in head 23-13 in GPT2 large are shown in Table 9. With the prediction "foo", the case is: 

\textcolor{orange}{Wall St. Bears Claw Back Into the Black (Reuters) Reuters - Short-sellers, Wall Street's dwindling band of ultra-cynics, are seeing green again.} : \textbf{\textcolor{red}{bar}} \textcolor{orange}{Stoking the Steamroller No other recording artist can channel American middle-class tastes quite like Chip Davis and his best-selling band.} : \textbf{bar} \textcolor{blue}{Liverpool completes signings of Alonso, Garcia LIVERPOOL, England (AP) -- Spanish pair Xabi Alonso from Real Sociedad and Luis Garcia from Barcelona signed five-year contracts with Liverpool on Friday.} : \textbf{\textcolor{orange}{foo}} \textcolor{blue}{U.S. Doping Watchdog to Question BALCO's Conte - IAAF HELSINKI (Reuters) - U.S . anti-doping officials plan to question Victor Conte after the BALCO head claimed he saw sprinter Marion Jones taking banned drugs, world athletics body the IAAF said Saturday.} : \textbf{foo} \textcolor{brown}{Liverpool Progresses to Champions League; Monaco, Inter Advance Four-time champion Liverpool progressed to soccer Champions League 2-1 on aggregate, overcoming a 1-0 home defeat to AK Graz in the second leg of qualifying.} \textbf{\textcolor{purple}{:}}

\begin{table}[htb]
\centering
\begin{small}
\begin{tabular}{lp{5.5cm}}
\toprule
\textbf{position} & \textbf{top words in vocabulary space} \\
\midrule
\textbf{\textcolor{red}{bar}}-value & \textbf{BAR}, \textbf{bars}, \textbf{Bars}, bart, \textbf{Bar}, bartender, \textbf{bar}, Barber \\
\textbf{bar}-value & bartender, \textbf{Bars}, bart, \textbf{bars}, \textbf{Bar}, Barber, \textbf{bar}, \textbf{BAR} \\
\textbf{\textcolor{orange}{foo}}-value & \textbf{foo}, McKenzie, \textbf{Foo}, Barney, Walters, Jenner, Murphy, lobster, Handler \\
\textbf{foo}-value & Walters, \textbf{foo}, Barney, McKenzie, Harrington, Murphy, Barber, Barron, Jenner \\
\textbf{\textcolor{red}{bar}}-key & Bloomberg, \textcolor{orange}{Investor, billionaires, CNBC, bankers, Companies, JPMorgan}, obal, \textcolor{orange}{economists}, bullish, Barron, \textcolor{orange}{HSBC, Friedman, Consumer, business, sellers}\\
\textbf{bar}-key & \textcolor{orange}{Buy}, Conn, Ok, Previous, Daily, NY, Yes, Anon, US, Ibid, \textcolor{orange}{Profit}, Staff, Journal, Vanguard, Tribune, \textcolor{orange}{Well}\\
\textbf{\textcolor{orange}{foo}}-key &  Buy, \textcolor{blue}{iverpool}, Ibid, \textcolor{blue}{YORK, UNITED}, Oliv, Charl, \textcolor{blue}{Location, Spanish, Miami, US, Liverpool}, \textcolor{purple}{Pool}, \textcolor{blue}{London, Greenwich, United}\\
\textbf{foo}-key & NYT, WATCH, Latest, Exclusive, Previous, UNC, \textcolor{blue}{US}, Watch, Possible, Ibid, Statement, Reaction, \textcolor{blue}{UK}, Reuters, \textcolor{blue}{United}, Smoke\\
\textbf{\textcolor{purple}{last}}-query & ruary, Pipe, lihood, swick, Flavoring, \textcolor{blue}{iverpool}, paddle, paraph, \textcolor{purple}{Lake}, Repe, tong, bole, etheless, \textcolor{purple}{Lakes} \\
\bottomrule
\end{tabular}
\end{small}
\caption{Top words of labels and last token in GPT2 large layer 23, head 13 on a sentence classification case.}
\end{table}

In this case, the false demonstrations with label "bar" are sampled from the "Business" class. The true demonstrations with label "foo" and the input text are sampled from the "Sports" class. On label positions' value-output vectors, "bar" and "foo" have top rankings. As for the key vectors at label positions, the labels correspond to business demonstrations extract the concepts about business, such as "investor" and "profit". The top tokens of true labels are related to places such as "Liverpool" and "Spanish", which exist in the corresponding demonstrations. These observations indicate that the value-output matrices extract label features, and the key matrix extract corresponding demonstration features. Analyzing the last position's query vector, we also observe concepts related to "Liverpool".

\section{Recency Datasets for Evaluation}
The three recency sentences transformed from the original sentence is shown in Table 10. 

\begin{table}[htb]
\centering
\begin{small}
\begin{tabular}{lp{5.5cm}}
\toprule
  & sentence \\
\midrule
origin & \textbf{S0 : bar S1 : bar S2 : foo S3 : foo S4 :}  \\
reorder-1 & \textbf{S2 : foo S0 : bar S1 : bar S3 : foo S4 :} \\
reorder-2 & \textbf{S0 : bar S2 : foo S3 : foo S1 : bar S4 :} \\
reverse & \textbf{S2 : foo S3 : foo S0 : bar S1 : bar S4 :}  \\
\bottomrule
\end{tabular}
\end{small}
\caption{Sentences transferred from origin sentence.}
\vspace{-10pt}
\end{table}

\end{document}